%% file: main.tex
\documentclass[sigconf]{acmart}

\AtBeginDocument{%
  \providecommand\BibTeX{{%
    \normalfont B\kern-0.5em{\scshape i\kern-0.25em b}\kern-0.8em\TeX}}}

\setcopyright{acmlicensed}
\copyrightyear{2018}
\acmYear{2018}
\acmDOI{XXXXXXX.XXXXXXX}

\acmConference[Conference acronym 'XX]{Make sure to enter the correct
  conference title from your rights confirmation emai}{June 03--05,
  2018}{Woodstock, NY}
\acmISBN{978-1-4503-XXXX-X/18/06}

\usepackage{multirow}
\usepackage{graphics}
\usepackage{enumitem}
\usepackage{array}
\usepackage{makecell}
\begin{document}

\title{Confidence-Aware Multi-Field Model Calibration}

\author{
Yuang Zhao$^{2*}$,
Chuhan Wu$^{1*}$,
Qinglin Jia$^{1*}$\authornote{Equal contribution.},
Hong Zhu$^4$,
Jia Yan$^3$,
Libin Zong$^3$,\\
Linxuan Zhang$^{2\dagger}$\authornote{Corresponding authors.},
Zhenhua Dong$^{1\dagger}$,
Muyu Zhang$^{3}$}
\renewcommand{\authors}{Yuang Zhao, Chuhan Wu, Qinglin Jia, Hong Zhu, Jia Yan, Libin Zong, Linxuan Zhang, Zhenhua Dong, Muyu Zhang}
\affiliation{%
  \institution{$^1$Noah's Ark Lab, Huawei \quad $^2$ Tsinghua University \\ $^3$Huawei Petal Cloud Technology Co., Ltd.\quad $^4$Consumer Cloud Service Interactive Media BU, Huawei}
  \country{}
  }
\email{zhaoya22@mails.tsinghua.edu.cn, lxzhang@tsinghua.edu.cn}
\email{{wuchuhan1, jiaqinglin2, zhuhong8, yanjia9,zonglibin,dongzhenhua,zhangmuyu}@huawei.com}

\renewcommand{\shortauthors}{Zhao and Wu, et al.}

\begin{abstract}
Accurately predicting the probabilities of user feedback, such as clicks and conversions, is critical for advertisement ranking and bidding.
However, there often exist unwanted mismatches between predicted probabilities and true likelihoods due to the rapid shift of data distributions and intrinsic model biases. 
Calibration aims to address this issue by post-processing model predictions, and field-aware calibration can adjust model output on different feature field values to satisfy fine-grained advertising demands.
Unfortunately, the observed samples corresponding to certain field values can be seriously limited to make confident calibrations, which may yield bias amplification and online disturbance.
In this paper, we propose a confidence-aware multi-field calibration method, which adaptively adjusts the calibration intensity based on confidence levels derived from sample statistics.
It also utilizes multiple fields for joint model calibration according to their importance to mitigate the impact of data sparsity on a single field.
Extensive offline and online experiments show the superiority of our method in boosting advertising performance and reducing prediction deviations.
\end{abstract}




\maketitle

\input{sections/introduction}

\input{sections/relatedwork.tex}
\input{sections/method.tex}

\input{sections/experiments.tex}
\input{sections/conclusion.tex}

\bibliographystyle{ACM-Reference-Format}
\bibliography{main}

\end{document}

%% file: sections/introduction.tex
\section{Introduction}

Estimating the real probabilities of user actions, such as click-through rate (CTR) and conversion rate (CVR), is a fundamental task in online advertising~\cite{user_action_prediction_richardson2007predicting, ctr_survey_chen2016deep, chapelle2014simple, tang2020progressive}.
In an ideal system, model-predicted probabilities are expected to reflect the true likelihoods of click and conversion, e.g., the real CTR should match the predicted CTR under sufficiently many ad impressions~\cite{niculescu2005predicting, lee2012estimating}.
Unfortunately, over- or under-estimation phenomena of model predictions are prevalent in real-world recommender systems for various reasons, including the limitation of binary targets~\cite{sir_deng2021calibrating, handbook_bella2010calibration}, model inductive biases~\cite{wenger2020non, overconfidence_bai2021don, gawlikowski2023survey}, and rapid shifts of online data distributions~\cite{data_shift_ovadia2019can, huang2021importance, shen2021towards, wang2022generalizing}.
This discrepancy often causes wasted advertising expenditure on ineffective ads, which damages the ROI of advertising campaigns and user experience~\cite{guo2022calibrated}.

\begin{figure}[!t]
  \centering
    \includegraphics[width=0.9\linewidth]{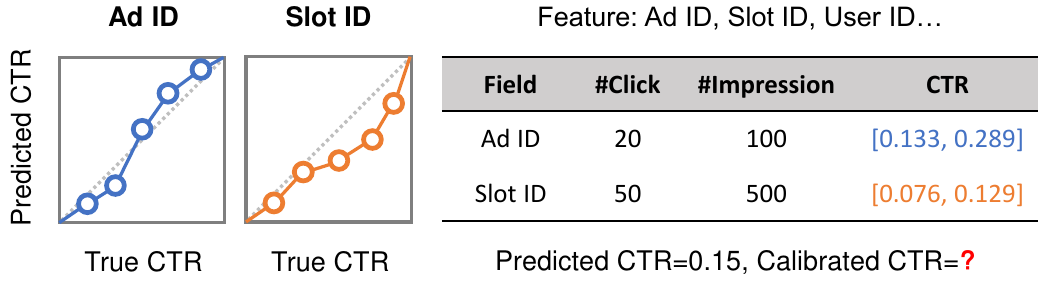}
  \vspace{-0.1in}
  \caption{An example of multi-field model calibration.}
  \vspace{-0.2in}
  \label{fig.exp}
\end{figure}

Model calibration post-processes the predicted scores based on the observation of real data to align the gaps between predicted and real probabilities.
Extensive studies have been conducted on global calibration, such as binning~\cite{histobin_zadrozny2001obtaining, bayesian_binning_naeini2015obtaining, isoreg_zadrozny2002transforming} and scaling~\cite{temp_scaling_guo2017calibration, platt_scaling_platt1999probabilistic, dirichlet_calib_kull2019beyond}, which adjust the predicted probabilities according to the posterior information of all observed samples.
However, global calibration suffers from the bias counteraction problem, e.g., the overall bias can be very small if scores for one advertiser are over-estimated while those for another advertiser are under-estimated~\cite{neucalib_pan2020field}.
Therefore, field-aware calibration is more practical in real-world scenarios, which can calibrate samples with the same value in a certain target field (e.g., ad ID) and alleviate biases in different sample subsets~\cite{adacalib_wei2022posterior}.

Pioneering studies like NeuCalib~\cite{neucalib_pan2020field} and AdaCalib~\cite{adacalib_wei2022posterior} have focused on field-aware calibration by learning additional calibration neural networks to adjust original model predictions.
However, samples corresponding to certain feature values are often too sparse to learn an unbiased and confident calibrating network.
As shown in Fig.~\ref{fig.exp}, the calculated CTR is 0.1 if an ad slot accumulates 50 clicks out of 500 impressions, while the 95\% confidence interval of CTR is [0.076, 0.129] (calculated by the Wilson interval formula~\cite{wilson_wilson1927probable}).
Such a wide range indicates strong uncertainty in field-aware calibration.
Furthermore, according to our industrial practice, the volumes of click and impression data w.r.t. most field values are smaller than the example above, which poses a severe challenge to the confidence of calibration models.
In addition, different fields may have diverse probability score distributions (Fig.~\ref{fig.exp}~left), and ad platforms may select multiple fields as the business targets.
Most existing methods mainly focus on single field calibration, and it is impractical to directly merge multiple fields due to the ultra sparsity of samples satisfying multiple field value conditions.

In this paper, we propose a confidence-aware multi-field calibration method named \textit{ConfCalib}, which can perform fine-grained calibration on multiple feature fields and meanwhile keep robustness under data sparsity via adaptive adjustment of the confidence level.
Concretely, we assume that user feedback follows a binomial distribution and use Wilson intervals to compute confidence-aware calibrated scores with a dynamic adjustment of calibration intensity. 
Moreover, we propose a simple yet effective multi-field fusion method to synthesize the calibration results on different fields, in order to further eliminate the influence of data sparsity.
Compared with methods based on neural networks, our ConfCalib performs more robustly in scenarios with severe data sparsity.
Extensive offline experiments in different datasets and online A/B test on our advertising platform demonstrate the effectiveness of our method.

The main contributions of this paper are summarized as follows:
\begin{itemize}[leftmargin=*]
    \item We propose a confidence-aware model calibration method to perform adaptive calibration on recommendation model predictions based on field-wise data sparsity.
    \item We introduce a simple yet effective joint calibration method on multiple fields to improve both calibration performance and robustness to data sparsity.
    \item Experiments on different offline datasets and the A/B testing on our advertising platform verify the superiority of our method.
\end{itemize}

%% file: sections/relatedwork.tex
\section{Related Work}
Model calibration is a classic problem in machine learning, which is proposed to eliminate the discrepancy between the predicted probabilities of classification models and the true likelihood~\cite{wald2021calibration, pleiss2017fairness, nixon2019measuring}.
It has been widely explored to promote accurate probability prediction in various applications, such as ad auctions~\cite{he2014practical, mcmahan2013ad}, weather forecasting~\cite{gneiting2005weather, degroot1983comparison}, and personalized medicine~\cite{jiang2012calibrating}.

Calibration methods calibrate trained prediction models during training or inference phase.
Methods integrated into training usually conduct calibration through well-designed loss functions, 
including AvUC~\cite{avuc_krishnan2020improving}, MMCE~\cite{mmce_kumar2018trainable}, Focal loss~\cite{focal_calib_mukhoti2020calibrating, focal_loss_lin2017focal} and SB-ECE~\cite{soft_calibration_objective_karandikar2021soft}, etc.
Besides, label smoothing, which can be interpreted as a modified primary loss, also serves to model calibration~\cite{label_smoothing_muller2019does, pereyra2017regularizing, liu2022devil}. 
This paper focus on post-hoc methods that calibrate model predictions by post-processing.
According to the characteristics of different mapping functions, existing calibration methods can be divided into three categories: scaling-based, binning-based, and hybrid methods.

\textbf{Scaling-based methods}~\cite{temp_scaling_guo2017calibration, attended_temp_scaling_mozafari2018attended, dirichlet_calib_kull2019beyond, platt_scaling_platt1999probabilistic, gamma_calib_kweon2022obtaining, beta_calib_kull2017beta, rahimi2020intra, gupta2021calibration} fit parametric mapping functions for the predicted scores. 
Platt Scaling~\cite{platt_scaling_platt1999probabilistic} adopts a linear mapping, assuming that the scores of each class follow normal distributions with the same variance. 
Beta calibration~\cite{beta_calib_kull2017beta}, Gaussian calibration~\cite{gamma_calib_kweon2022obtaining} and Gamma calibration~\cite{gamma_calib_kweon2022obtaining} further assume more complicated data distributions and derive corresponding score mapping functions to adapt more general situations. 
Dirichlet calibration~\cite{dirichlet_calib_kull2019beyond} generalizes Beta calibration to the multi-class classification by assuming a Dirichlet distribution.
The above methods are limited to a few occasions due to the specific distribution assumptions. 
Temperature Scaling~\cite{temp_scaling_guo2017calibration} considers that the miscalibration of models comes from the overfitting of negative log-likelihood (NLL) during training, then proposes a temperature multiplier for logits of all classes to reduce test NLL and eliminate miscalibration.
Although these methods have solid theoretical validity, they are difficult to deal with complex practical scenarios.

\textbf{Binning-based methods}~\cite{isoreg_zadrozny2002transforming, isoreg_new_menon2012predicting, bayesian_isoreg_neelon2004bayesian,  histobin_zadrozny2001obtaining, bayesian_binning_naeini2015obtaining} divide samples into several bins after sorting them by predicted scores. 
Histogram Binning~\cite{histobin_zadrozny2001obtaining} adopts an equal-frequency binning and directly uses the mean positive rate in each bin as the calibrated scores.
Bayesian Binning~\cite{bayesian_binning_naeini2015obtaining} considers multiple binning schemes and combines them using a derived Bayesian score to yield more robust calibrated predictions. 
To maintain the original order of model predictions, Isotonic Regression~\cite{isoreg_zadrozny2002transforming} adjusts the binning bounds by minimizing the residual between calibrated predictions and labels, hence ensuring monotonicity of the posterior probability in bins.

\textbf{Hybrid methods}~\cite{scaling_binning_kumar2019verified, mix_n_match_zhang2020mix, sir_deng2021calibrating, MBCT_huang2022mbct, neucalib_pan2020field, adacalib_wei2022posterior} combine scaling and binning methods, which are superior in calibration performance and prevalence in industry.
Scaling-binning~\cite{scaling_binning_kumar2019verified} first fits a mapping function for the sorted samples and then employs histogram binning calibration on the mapped scores. 
Smoothed Isotonic Regression (SIR)~\cite{sir_deng2021calibrating} uses isotonic regression on bins and fits a monotonic piece-wise linear calibration function. 
Recently, some works have focused on field-wise calibration, which aims to reduce the calibration error in a specific feature field.
Neural Calibration~\cite{neucalib_pan2020field} learns a neural isotonic line-plot scaling function and an auxiliary network directly taking features as input to capture feature field information.
AdaCalib~\cite{adacalib_wei2022posterior} extends the single global mapping function of Neural Calibration to different ones for each value of a certain feature field and dynamically adjusts the number of bins by neural networks.
These neural methods may suffer from data sparsity in specific application scenarios where network training is hard to converge.
Additionally, existing field-wise calibration methods simply consider one field and ignore potential relations among fields, meaning there is still a lack of research on multi-field calibration methods.

%% file: sections/method.tex
\section{Methodology}\label{sec:Model}

\begin{figure*}[!t]
  \centering
    \includegraphics[width=0.85\linewidth]{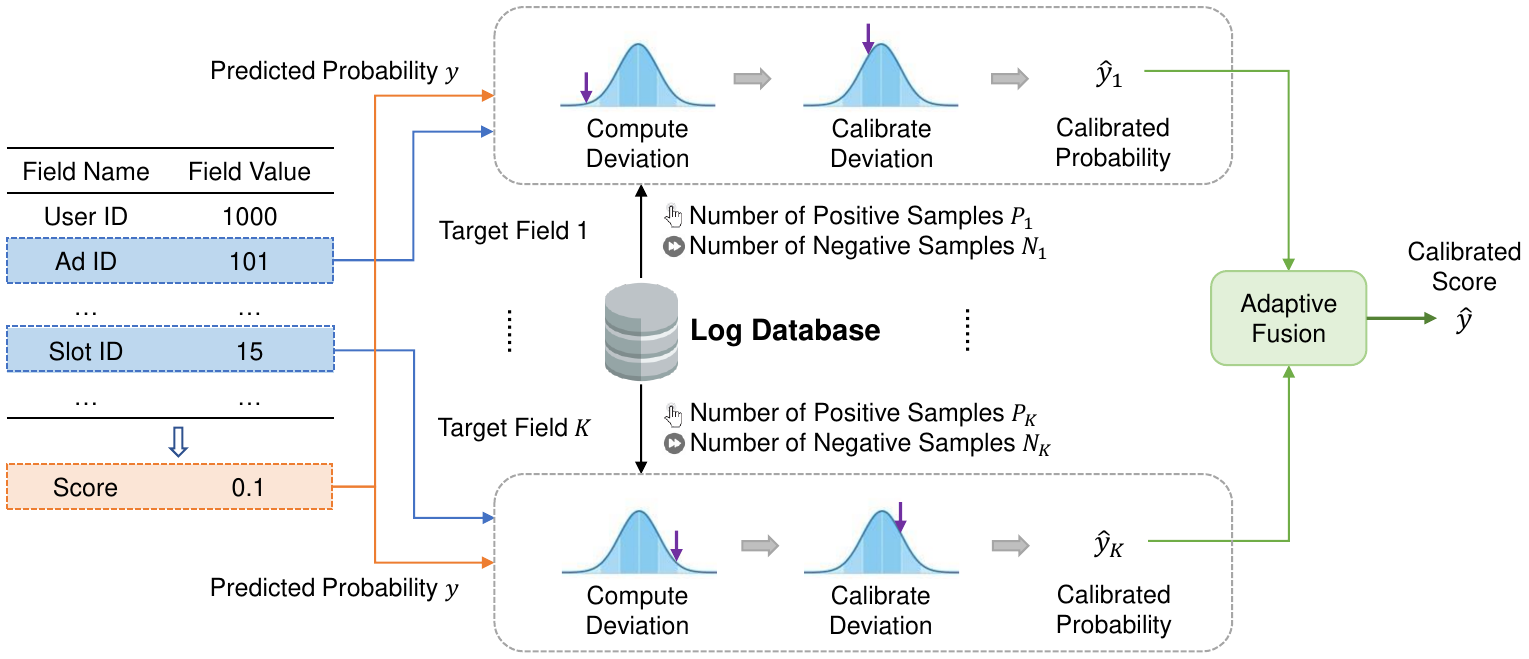}
  \vspace{-0.1in}

  \caption{The overall framework of our \textit{ConfCalib} method.}
  \vspace{-0.1in}
  \label{fig.model}
\end{figure*}

The overall framework of our multi-field confidence-aware calibration method is shown in Fig.~\ref{fig.model}.
It first performs confidence-aware calibration on multiple target feature fields independently and then fuses the scores calibrated on different feature fields into a unified one.
The details of our method are introduced as follows.

\subsection{Problem Definition}

First, we briefly introduce the problem setting in this paper, namely field-level calibration on the binary classification problem.
Suppose we have a trained binary classification model $f(\cdot)$ to provide predictions $\hat{p}=f(x)$ on a dataset $\{x_i, y_i\}_{i=0}^{N}$, where $x_i$ denotes the features with multiple fields and $y_i$ denotes the binary label. 
We generally consider biases on the level of sample subsets since the true likelihood of a specific sample is unobservable.
For a specific sample subset, the average predicted probability usually deviates from the observed positive sample rate, thus requiring calibrations.
Assume there are $K$ target feature fields and the value of each field is denoted as $z_j~(j=1,2,\dots,K)$, the calibration model fits a transformation function $g(\cdot;z_1,\dots,z_K)$ and gives calibrated probability  $\hat{p}_{\text{calib}}=g(\hat{p};z_1,\dots,z_K)$ according to sample statistics on target fields.
The goal of field-level calibration on a certain field is to minimize the field-level calibration error, such as Field-RCE~\cite{neucalib_pan2020field}.
For multi-field calibration, we can employ metrics like ECE~\cite{bayesian_binning_naeini2015obtaining} and MVCE~\cite{MBCT_huang2022mbct} to measure the global calibration error. The above metrics will be detailed in Section~\ref{sec:Metrics}.

\begin{figure*}[!t]
  \centering
    \includegraphics[width=0.87\linewidth]{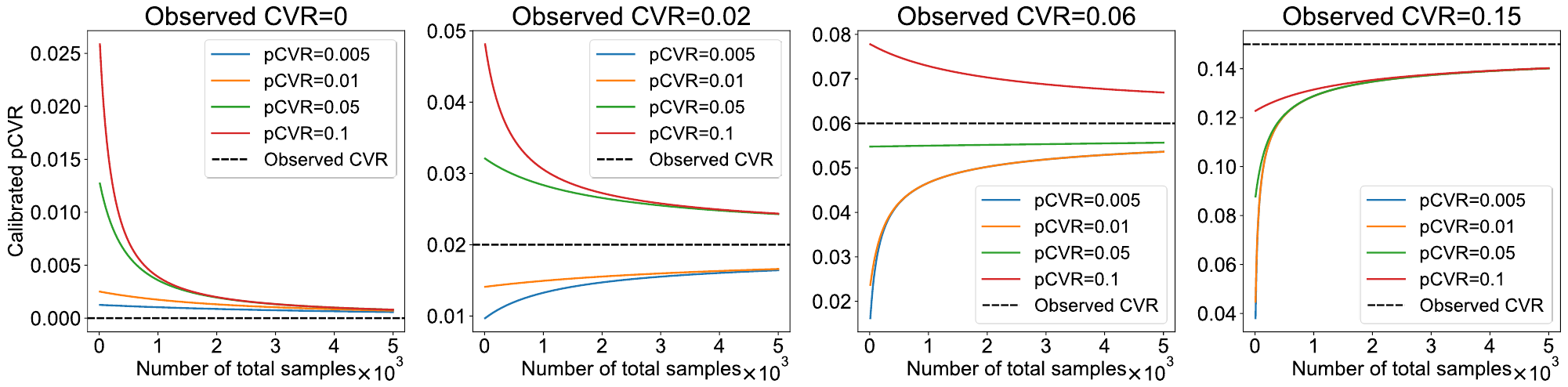}
      \vspace{-0.1in}

  \caption{The curves of calibrated scores w.r.t. different predicted scores under different numbers of observed samples.}
    \vspace{-0.1in}

  \label{fig.calibration_curve}
\end{figure*}
\subsection{Confidence-aware Calibration}\label{sec:confcalib}

Next, we introduce our confidence-aware calibration method.
In practical scenarios, field-level calibration faces severe data sparsity issues due to the limited time window for data collection and intrinsic sparsity of certain user feedback, such as conversions.
Consequently, calibrations on feature fields with numerous values are neither accurate nor confident.
The main problem caused by data sparsity is that the observed probability is not credible enough to reflect the true data distribution.
In this case, complete trust in limited observations will result in overfitting and harm generalization.
In summary, it is necessary to preserve some trust in model predictions while referencing observations under data sparsity.

Most user feedback, like clicks and conversions, can be naturally modeled using binomial distributions.
Given the observed positive sample ratio $p$ and the number of total samples $n$, we can use a confidence interval formula to compute the confidence interval (CI) ($p_-$, $p_+$) of the real feedback probability under a certain confidence level. Conversely, given a certain predicted feedback probability, we can obtain the corresponding confidence level by regarding the predicted value as the bound of CI. In other words, there is a mapping between the predicted probability and the confidence level, which is the core of our confidence-aware calibration method.

The most commonly used CI formula is the normal interval. However, due to the limitation of the central limit theorem, it is unreliable with a small sample size or an extreme probability close to 0 or 1, which is quite common for user feedback prediction in recommender systems. 
Therefore, we adopt the \textbf{Wilson confidence interval}~\cite{wilson_wilson1927probable}, which can be safely employed under the special circumstances above. The formula is
\begin{equation}\small
    (p_-, p_+)\approx {\frac {1}{~1+{\frac {\,z^{2}\,}{n}}~}}\left({p}+{\frac {\,z^{2}\,}{2n}}\right)~\pm ~{\frac {z}{~1+{\frac {z^{2}}{n}}~}}{\sqrt {{\frac {\,{p}(1-{p})\,}{n}}+{\frac {\,z^{2}\,}{4n^{2}}}~}},\label{interval}
\end{equation}
where $z$ is the deviation score (e.g., $z=1.96$ for 95\% confidence).

In order to perform field-level calibration, our method operates at the level of sample subsets, which can be obtained by splitting samples based on different values of the target feature field.
Besides, field-wise binning\footnote{Binning is optional in our method since it introduces heavier data sparsity.}, i.e., further dividing each subset into several bins by order of their predicted scores, is optional.
For each sample subset, we need the statistics,
including the number of samples $n$, the ratio of positive samples $p$, and the average predicted score $\hat{p}$.
Regard $\hat{p}$ as a certain bound\footnote{Let $p_+=\hat{p}$ if $\hat{p}>p$, else $p_-=\hat{p}$.} of CI and substitute these variables into Eq.~(\ref{interval}), we can solve the deviation score $z\in[0, +\infty)$ using numerical algorithms, such as the bisection method.
A larger $z$ indicates a greater bias between model predictions and current observations, but a higher confidence level as well.
We can directly use $p$ as the calibrated score, i.e., change $z$ to $0$, to entirely eliminate biases.
However, this will minimize the confidence level and result in poor generalization under data sparsity.
Therefore, we intend to make the calibrated score between original $p$ and predicted $\hat{p}$, thus corresponding to a relatively high confidence. 
To this end, we propose a non-linear transformation function $g(z)$ to reduce the deviation.
Intuitively, $g(z)$ should satisfy the following requirements: 
(1) passing through the origin: $g(0)=0$; (2) monotonic: $\forall z_1 \leq z_2, g(z_1)\leq g(z_2)$;  (3) bounded: $\exists C\in\mathbb{R}^+, g(+\infty)\leq C$.
Thus, we devise a variant of the sigmoid function as follows:
\begin{equation}\small
    \label{eq.z_map}
    g(z) = \lambda\cdot\left(\frac{2}{1+e^{-z/2}}-1\right), z\in[0, +\infty),
\end{equation}
where $\lambda$ is a hyperparameter used to provide an upper bound and control the calibration intensity manually.
A larger calibration intensity can result in a smaller $z$, meaning more belief in observations and less belief in original predictions.
Obviously, we have $g(+\infty)=\lambda$ so that a large deviation is calibrated to a small one (e.g., the deviation corresponding to 95\% confidence level if $\lambda=1.96$), and minor deviations are reduced almost linearly.
By replacing the original $z$ in Eq.~(\ref{interval}) with the transformed value $g(z)$, we can obtain a new CI, then one of its bounds is the desired confidence-aware calibrated score $\hat{p}'$.
In this way, given a test sample located in the same subset, we compute a field-wise calibrated score $p_\text{calib}$ by scaling its original predicted probability $p_\text{pred}$, i.e., 
\begin{equation}\small
    p_\text{calib} = \frac{\hat{p}'}{\hat{p}}p_\text{pred}.
\end{equation}

Fig.~\ref{fig.calibration_curve} shows the curves of calibrated probabilities w.r.t variant predicted probabilities under different observed sample numbers and positive sample rates, taking the conversion rate prediction task as an example.
We uniformly set $\lambda=2$ for the z-score transformation function. 
The curves demonstrate that the predicted probability is calibrated towards the observed one, and the proximity to the observation increases as the number of samples increases.

Overall, our confidence-aware method adjusts model-predicted probabilities by resetting the confidence level, which considers the observation and prediction together to avoid violent value changes and overfitting limited observations.
Consequently, it keeps robust under severe data sparsity, especially circumstances with few observed samples and zero positive samples.

\subsection{Multi-field Joint Calibration}
Previous field-level calibration methods only calibrate and evaluate errors in a certain field, ignoring other fields. 
However, the field-level calibration error in different fields may vary greatly, which is ignored by single-field calibration.
In addition, the biases in different fields, as reflections of the global bias from different perspectives, have a certain degree of commonality. 
Therefore, data sparsity faced by a certain field is expected to be alleviated through calibrations on other fields.

A naive way to extend calibration from a single field to multiple fields is to split sample subsets based on the combination of feature values of all the fields. 
Nevertheless, it will cause severer data sparsity since fewer samples satisfy multiple field value conditions.
Accordingly, prior single-field calibration methods cannot be directly extended to multi-field calibration.
In contrast, our method demonstrates a better extensibility among fields.
The confidence-aware calibration in a single field in Section~\ref{sec:confcalib} is actually a scaling operation for a specific sample, thus allowing a generalization to multi-field joint calibration.
Assume that there are $K$ target fields to calibrate. 
For each sample, we can obtain $K$ scaling multipliers $m_1, m_2, \dots, m_K$, where $m_i={\hat{p_i}'}/{\hat{p_i}}$ for the $i$-th field.
We adaptively fuse these multipliers via the weighted geometric mean, that is
\begin{equation}
    p_\text{calib}=m_1^{w_1}m_2^{w_2}\cdots m_K^{w_K}\cdot p_\text{pred},
\end{equation}
where $w_1, w_2, \dots, w_K$ is weights satisfying $\sum_{i=1}^K{w_i}=1$. 
The fusion weights can be precisely obtained by grid search.
The proposed fusion method allows us to conduct joint calibration based on the marginal distribution of each field independently instead of dealing with the joint distribution, thus avoiding further sparsity of samples.

\begin{figure}[!b]
  \vspace{-0.1in}
  \centering
    \includegraphics[width=0.99\linewidth]{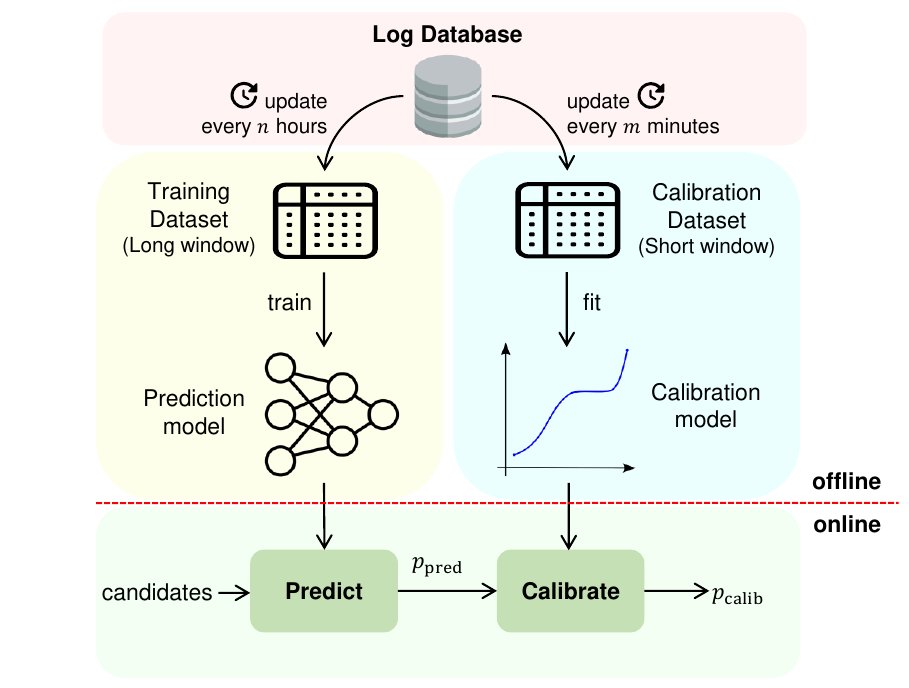}
  \vspace{-0.1in}

  \caption{Online deployment diagram of ConfCalib.}
  \label{fig.deployment}
\end{figure}

\subsection{Online Deployment}
Fig.~\ref{fig.deployment} shows the online deployment diagram of ConfCalib in our advertising platform. 
On the offline side, the prediction model's training and ConfCalib's fitting are paralleled and completely independent.
At fixed intervals, the calibration module pulls the latest online samples from the log database and refits calibration mappings.
This interval is usually set relatively shorter (e.g., 30 minutes) than the prediction model update period to better capture the dynamic shift of online data.
On the online side, prediction and calibration are conducted in sequence. 
First, the prediction model makes predictions for candidate objects, then the calibration model calibrates the predicted probabilities and outputs the calibrated probabilities for subsequent use.

Recent methods designed for advertising platforms, such as NeuCalib and AdaCalib, involve auxiliary neural networks.
These calibration models are coupled with original prediction models during the training phase, thus increasing the complexity of online deployment. 
On the contrary, our ConfCalib relies entirely on posterior probabilities of online samples and fits parameters without training networks.
Therefore, it can be directly inserted into the pipeline without interfering with the training stage of prediction models, resulting in lower deployment costs.

%% file: sections/experiments.tex
\section{Experiments}

\subsection{Experimental Setup}
\subsubsection{Settings}

The pipeline of our experiment is as follows.
Each dataset is conventionally split into three parts in chronological order, i.e., training, validation, and test set.
Firstly, we train a base neural network on the training set as the prediction model and fix it.
Then, the base model makes predictions for validation set samples to fit (or train) calibration models.
Finally, we use the test set to evaluate the performance of calibration methods.

For all the datasets and methods, the prediction model is a fully connected network with four layers.
Hyperparameters of baselines are set according to the original papers or carefully tuned.
Results of neural methods are the average results of five random experiments with means and standard deviations reported.
For our ConfCalib, we set the number of bins as \{50, 10\} and $\lambda$ in Eq.~(\ref{eq.z_map}) as \{1.0, 1.6\} for Avazu and our industrial dataset.

\begin{table}[!h]
\small
    \vspace{-0.05in}
  \caption{Data statistics of the public and industrial datasets.}
    \vspace{-0.1in}
  \label{tab:data_stats}
  \begin{tabular}{lcccc}
    \toprule
    \textbf{Dataset} & \textbf{Task} & \textbf{\#Training} & \textbf{\#Validation} & \textbf{\#Test}\\
    \midrule
    Avazu & CTR & 28.3M & 4.04M & 8.08M\\
    Industrial & CVR & 27.5M & 3.8M & 7.5M \\
  \bottomrule
\end{tabular}
  \vspace{-0.15in}

\end{table}

\subsubsection{Datasets}
We conduct experiments on our industrial dataset and a public dataset Avazu~\footnote{https://kaggle.com/competitions/avazu-ctr-prediction}.
Data statistics are shown in Table~\ref{tab:data_stats}.
\textbf{Avazu} is a popular dataset on mobile ad CTR prediction with 40.4 million samples in 10 days ordered by the impression time. 
We split the dataset by day in a ratio of 7:1:2 for training, validation, and test. 
For our multi-field joint calibration method, the target fields are set to "site\_id" (3480)\footnote{Represent the number of possible values of the corresponding field.}, "app\_id" (4864) and "app\_domain" (305). 
For other methods considering a single field, the target field is set to "site\_id".
The \textbf{industrial} dataset is collected from the click logs on our ad platform in 10 consecutive days, and the prediction target is conversion, with a split ratio of 7:1:2 by day. 
For our method, the target fields are "advertiser\_id" (529), "app\_size" (82), and "app\_id" (56). For other methods, the target field is "advertiser\_id".
The ratios of validation sets are relatively small while splitting, which is more in line with the short time window of online data collection.

\subsubsection{Baselines}
We choose several representative methods of the three categories as baselines for comparison.
The scaling-based methods include naive scaling (multiply all predicted scores by a value $k$ to make their mean equal to the positive sample ratio), Platt Scaling (PlattScaling)~\cite{platt_scaling_platt1999probabilistic}, Gaussian Calibration (GaussCalib)~\cite{gamma_calib_kweon2022obtaining} and Gamma Calibration (GammaCalib)~\cite{gamma_calib_kweon2022obtaining}.
The binning-based methods include Histogram Binning (HistoBin)~\cite{histobin_zadrozny2001obtaining} and Isotonic Regression (IsoReg)~\cite{isoreg_zadrozny2002transforming}. 
The hybrid methods include Smoothed Isotonic Regression (SIR)~\cite{sir_deng2021calibrating}, Neural Calibration (NeuCalib)~\cite{neucalib_pan2020field} and AdaCalib~\cite{adacalib_wei2022posterior}, with the latter two methods based on neural networks. 
The number of bins for NeuCalib and AdaCalib is set according to the initial settings when proposed. 

\subsubsection{Evaluation Metrics}\label{sec:Metrics}
Referring to prior methods, we continue to use field-level relative calibration error (Field-RCE or F-RCE)~\cite{neucalib_pan2020field} to evaluate the calibration error on a certain field, which formula is
\begin{equation}\small
    \text{Field-RCE}=\frac{1}{|\mathcal{D}|}\sum_{z\in\mathcal{Z}}{|\mathcal{D}_z|\frac{\left|\sum_{(x_i,y_i)\in D_z}{(y_i-\hat{p}_i)}\right|}{\sum_{(x_i,y_i)\in D_z}{y_i}}},
\end{equation}
where $\mathcal{D}$ is the dataset, $\mathcal{Z}$ is the possible value set of the target field and $\mathcal{D}_z$ is the subset of samples with the target field value $z$ satisfying $\sum_{z\in\mathcal{Z}}{|\mathcal{D}_z|}=|\mathcal{D}|$. 
Intuitively, a lower Field-RCE indicates lower biases along the target field.
Furthermore, we use the mean Field-RCE on multiple fields (Multi-Field-RCE or MF-RCE) to measure the multi-field calibration performance.

Meanwhile, we adopt expected calibration error (ECE)~\cite{bayesian_binning_naeini2015obtaining} and multi-view calibration error (MVCE)~\cite{MBCT_huang2022mbct} to evaluate the global calibration error. 
ECE arranges all predicted values in order and divides them equally into several subsets, i.e., a binning operation. Then it is calculated as
\begin{equation}\small
    \text{ECE}=\frac{1}{|\mathcal{D}|}\sum_{t=1}^T{{\left|\sum\nolimits_{(x_i,y_i)\in \mathcal{D}_{t}}{(y_i-\hat{p}_i)}\right|}},
\end{equation}
where $T$ is the number of bins and $\mathcal{D}_t$ is the subset of samples in the $t$-th bin.
MVCE iteratively shuffles the dataset and calculates ECE until the accumulative mean ECE converges. The formula is
\begin{equation}\small
    \text{MVCE}=\frac{1}{RT}\sum_{r=1}^R\sum_{t=1}^T{\frac{\left|\sum_{(x_i,y_i)\in \mathcal{D}_{r,t}}{(y_i-\hat{p}_i)}\right|}{|\mathcal{D}_{r,t}|}},
\end{equation}
where $R$ is the number of shuffle operations, $T$ is the number of bins, and $\mathcal{D}_{r,t}$ is the subset of samples in the $t$-th bin after the $r$-th shuffle. 
MVCE allows us to inspect the calibration error from a more global perspective beyond the level of feature fields, thereby highlighting the effectiveness of multi-field calibration.

Additionally, we also report AUC and LogLoss to evaluate the ranking performance after calibration since our method and primary baselines are not globally isotonic.

\begin{table*}[!t]
  \small
  \caption{The performance of different methods in terms of prediction accuracy and miscalibration degrees. ``$\uparrow$'' means higher is better, ``$\downarrow$'' means lower is better.}
    \vspace{-0.1in}
  \label{tab:main_results}
  \resizebox{0.9\linewidth}{!}{
  \begin{tabular}{c|c|lcccccc}
    \toprule
    \textbf{Dataset} & \textbf{Type} & \textbf{Method} & \textbf{AUC$\uparrow$} & \textbf{LogLoss$\downarrow$} & \textbf{Field-RCE$\downarrow$} & \textbf{Multi-Field-RCE$\downarrow$} & \textbf{ECE$\downarrow$} & \textbf{MVCE$\downarrow$} \\

    \midrule[0.08em]
    \multirow{11}{*}{\makecell[c]{\textbf{Avazu}\\(CTR)}} & No Calib. & N/A 
     & 0.7545 & 0.3718 & 0.2295 & 0.1837 & 0.0123 & 0.0117\\
    \cline{2-9}
    \multirow{11}{*}{} & \multirow{2}{*}{Binning} & HistoBin~\cite{histobin_zadrozny2001obtaining}  & 0.7543 & 0.3714 & 0.2034 & 0.1457 & 0.0149 & \underline{0.0038}\\
    \multirow{11}{*}{} & \multirow{2}{*}{} & IsoReg~\cite{isoreg_zadrozny2002transforming}  & 0.7545 & 0.3713 & 0.1997 & 0.1432 & 0.0082 & 0.0039\\
    \cline{2-9}
    \multirow{11}{*}{} & \multirow{4}{*}{Scaling} & Naive  & 0.7545 & 0.3713 & 0.1874 & 0.1381 & \underline{0.0056} & \underline{0.0038}\\
    \multirow{11}{*}{} & \multirow{4}{*}{} & PlattScaling~\cite{platt_scaling_platt1999probabilistic}  & 0.7545 & 0.3712 & 0.1868 & 0.1383 & 0.0058 & 0.0039\\
    \multirow{11}{*}{} & \multirow{4}{*}{} & GaussCalib~\cite{gamma_calib_kweon2022obtaining}  & 0.7545 & 0.3712 & 0.1969 & 0.1470 & 0.0058 & 0.0041\\  
    \multirow{11}{*}{} & \multirow{4}{*}{} & GammaCalib~\cite{gamma_calib_kweon2022obtaining}  & 0.7545 & 0.3712 & 0.1984 & 0.1480 & 0.0058 & 0.0041\\
    \cline{2-9}
    \multirow{11}{*}{} & \multirow{4}{*}{Hybrid} & SIR~\cite{sir_deng2021calibrating}  & 0.7545 & 0.3717 & 0.2316 & 0.1849 & 0.0122 & 0.0116\\
    \multirow{11}{*}{} & \multirow{4}{*}{} & NeuCalib~\cite{neucalib_pan2020field} & 0.7554$\pm$0.0008 & 0.3709$\pm$0.0006 & \underline{0.1491$\pm$0.0209} & \underline{0.1091$\pm$0.0225} & 0.0110$\pm$0.0023 & 0.0042$\pm$0.0021\\
    \multirow{11}{*}{} & \multirow{4}{*}{} & AdaCalib~\cite{adacalib_wei2022posterior}  & \underline{0.7569$\pm$0.0003} & \underline{0.3696$\pm$0.0002} & 0.1650$\pm$0.0155 & 0.1308$\pm$0.0247 & 0.0099$\pm$0.0012 & 0.0065$\pm$0.0029\\
    \multirow{11}{*}{} & \multirow{4}{*}{} & ConfCalib 
     & \textbf{0.7575} & \textbf{0.3693} & \textbf{0.1435} & \textbf{0.0967} & \textbf{0.0037} & \textbf{0.0015}\\
    
    \midrule[0.08em]
    \multirow{13}{*}{\makecell[c]{\textbf{Industrial}\\(CVR)}} & No Calib. & N/A  & 0.8226 & 0.0868 & 0.3657 & 0.3015 & 0.0022 & 0.0008 \\
    \cline{2-9}
    \multirow{11}{*}{} & \multirow{2}{*}{Binning} & HistoBin & 0.8223 & 0.0868 & 0.3838 & 0.3145 & 0.0017 & 0.0015 \\
    \multirow{11}{*}{} & \multirow{2}{*}{} & IsoReg & 0.8226 & 0.0867 & 0.3779 & 0.3079 & 0.0036 & 0.0015 \\
    \cline{2-9}
    \multirow{11}{*}{} & \multirow{4}{*}{Scaling} & Naive & 0.8226 & 0.0868 & 0.3555 & 0.2811 & 0.0020 & 0.0016 \\
    \multirow{11}{*}{} & \multirow{4}{*}{} & PlattScaling & 0.8226 & 0.0868 & 0.3929 & 0.3190 & 0.0018 & 0.0015 \\
    \multirow{11}{*}{} & \multirow{4}{*}{} & GaussCalib & 0.8226 & 0.0867 & 0.3606 & 0.2916 & 0.0017 & 0.0015\\  
    \multirow{11}{*}{} & \multirow{4}{*}{} & GammaCalib & 0.8226 & 0.0867 & 0.3756 & 0.3047 & 0.0018 & 0.0015 \\
    \cline{2-9}
    \multirow{11}{*}{} & \multirow{4}{*}{Hybrid} & SIR & 0.8226 & 0.0868 & 0.3555 & 0.2811 & 0.0020 & 0.0016 \\
    \multirow{11}{*}{} & \multirow{4}{*}{} & NeuCalib & \underline{0.8271$\pm$0.0020} & \underline{0.0866$\pm$0.0002} & \underline{0.2856$\pm$0.0366} & 0.2330$\pm$0.0234 & 0.0012$\pm$0.0006 & 0.0007$\pm$0.0004 \\
    \multirow{11}{*}{} & \multirow{4}{*}{} & AdaCalib & 0.8192$\pm$0.0005 & 0.0875$\pm$0.0001 & 0.3234$\pm$0.0231 & \underline{0.2307$\pm$0.0295} & \underline{0.0010$\pm$0.0003} & 0.0007$\pm$0.0004 \\
    \multirow{11}{*}{} & \multirow{4}{*}{} & ConfCalib & \textbf{0.8303} & \textbf{0.0860} & \textbf{0.2728} & \textbf{0.2243} & \textbf{0.0007} & 0.0007 \\
    \bottomrule
  \end{tabular}
  }
\end{table*}

\begin{figure*}[!h]
  \centering
    \includegraphics[width=0.9\linewidth]{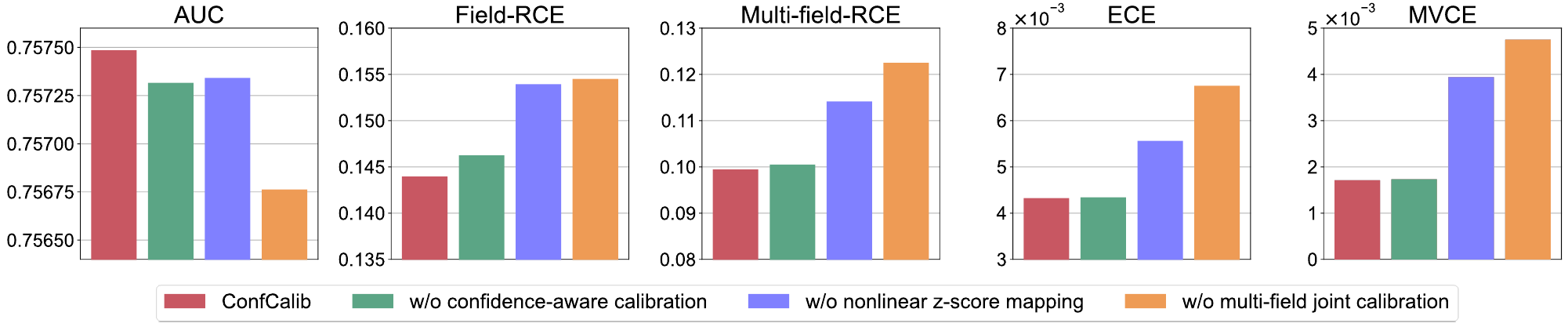}
      \vspace{-0.15in}
  \caption{Results of the ablation study on several variants of ConfCalib.}
    \vspace{-0.1in}
  \label{fig.ablation}
\end{figure*}

\subsection{Results}

\subsubsection{Main Results}
The main results on each dataset are shown in Table~\ref{tab:main_results}. 
Our method shows better calibration performance compared with baselines.
It achieves lower calibration errors on both field-wise and global metrics compared to the uncalibrated model and almost all the baselines.
Besides, our method also achieves higher AUC, indicating an improvement in ranking ability.
Scaling and binning-based baselines show poor performances.
They reduce global errors to some extent through global calibration while powerless against field-wise calibration errors.
Field-level calibration methods such as NeuCalib and AdaCalib achieve low calibration errors on the single target field among baselines. 
However, they are still inferior to our method, especially on multi-field metrics like Multi-Field-RCE and global metrics like ECE, demonstrating the effectiveness of our multi-field joint calibration.
The two neural methods show large variance, which means unstable learning on sparse data. 
On the contrary, our method only relies on posterior statistical information and maintains a stable calibration effect.
Moreover, AdaCalib obtains a lower AUC than the uncalibrated model on our industrial dataset, indicating that complex neural calibration methods may be counterproductive under data sparsity.

\subsubsection{Ablation Study}
We report the results of several variants of our ConfCalib on Avazu CTR task in Fig.~\ref{fig.ablation}. 
Concretely, we respectively remove the confidence-aware calibration (i.e., turn into a field-wise naive calibration), the nonlinear z-score mapping function (replaced by a single linear mapping), and the multi-field joint calibration from the original ConfCalib.
The removal of our confidence-aware calibration results in increase in field-wise calibration errors and decrease in AUC.
After replacing the mapping function with a single linear function $g(z)=\lambda z$, AUC and all calibration error metrics deteriorate.
This validates that the bounded mapping can provide proper calibration intensity compared with simple scaling. 
Finally, removing the joint calibration on multiple fields significantly increases both field-wise and global calibration errors and causes a large decline in AUC.

\begin{figure*}[!h]
  \centering
  \vspace{-0.1in}
    \includegraphics[width=0.9\linewidth]{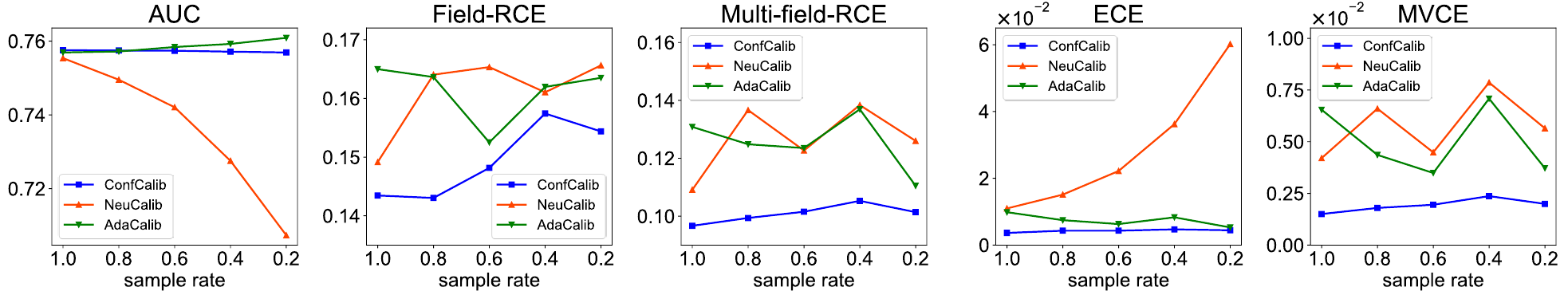}
    \vspace{-0.1in}

  \caption{Results of robustness analysis of different data sparsity levels.}
    \vspace{-0.1in}

  \label{fig.robustness}
\end{figure*}

\begin{table}[!h]
  \caption{Results of recalibration on learning-based baselines.}
\vspace{-0.15in}
  \label{tab:recalib}
  \small
  \resizebox{0.85\columnwidth}{!}{%
  \begin{tabular}{cccccc}
    \toprule
    \textbf{Method} & \textbf{AUC$\uparrow$} & \textbf{F-RCE$\downarrow$} & \textbf{MF-RCE$\downarrow$} & \textbf{ECE$\downarrow$} & \textbf{MVCE$\downarrow$}\\
    \midrule
    GaussCalib & 0.7545 & 0.1969 & 0.1470 & 0.0058 & 0.0041\\  
    +ConfCalib  & \textbf{0.7577} & \textbf{0.1417} & \textbf{0.1105} & \textbf{0.0053} & 0.0050\\\midrule
    GammaCalib & 0.7545 & 0.1984 & 0.1480 & 0.0058 & 0.0041\\
   +ConfCalib  & \textbf{0.7577} & \textbf{0.1430} & \textbf{0.1113} & \textbf{0.0053} & 0.0050\\\midrule
    NeuCalib  & 0.7554  & 0.1491 & 0.1091 & 0.0110 & 0.0042 \\
   +ConfCalib  & \textbf{0.7558} & \textbf{0.1446} & \textbf{0.0994} & \textbf{0.0099} & \textbf{0.0030}\\\midrule
    AdaCalib  & 0.7569  & 0.1650 & 0.1308 & 0.0099 & 0.0065 \\
   +ConfCalib  & 0.7569 & \textbf{0.1535} & \textbf{0.1069} & \textbf{0.0072} & \textbf{0.0024}\\
  \bottomrule
\end{tabular}
}
  \vspace{-0.15in}
\end{table}

\subsection{Recalibration on Learning-Based Baselines}
Part of the baselines are learning-based, such as GaussCalib, GammaCalib, NeuCalib, and AdaCalib. 
Since our ConfCalib does not require extra model training and has a low deployment cost, it can recalibrate the calibrated output of other methods for further improvement. 
We experiment on Avazu CTR task.
First, we fit our method and baselines to be recalibrated separately on the validation set.
Then, baselines make calibrated predictions for the test set.
Finally, our method conducts recalibration on the calibrated outputs.
Table~\ref{tab:recalib} shows that recalibration achieves further improvement compared with the original results of other methods.
Profiting from multi-field joint calibration, recalibration using ConfCalib significantly reduces Multi-Field-RCE and the global calibration error MVCE. 
Besides, recalibration on methods that strictly keep the order of original predictions, such as GaussCalib and GammaCalib, can provide an improvement in AUC as well.

\subsection{Robustness Analysis against Data Sparsity} 
One primary advantage of ConfCalib is its robustness against data sparsity. 
Due to the limited time window for online data collection, samples available to fit calibration models can be very few.
Under such circumstances, methods based on neural networks may be ineffective or even counterproductive.
Here, we reduce the validation dataset used for calibration in Avazu with various sample rates to simulate scenarios with varying degrees of data sparsity. 
The calibration performances w.r.t different sparsity degrees are shown in Fig.~\ref{fig.robustness}.
As the number of samples for fitting the calibration model decreases, the performances of NeuCalib, AdaCalib get worse evidently, while our ConfCalib is little affected and exhibits strong robustness against data sparsity.

\begin{figure}[!t]
    \vspace{-0.1in}
  \centering
    \includegraphics[width=0.725\linewidth]{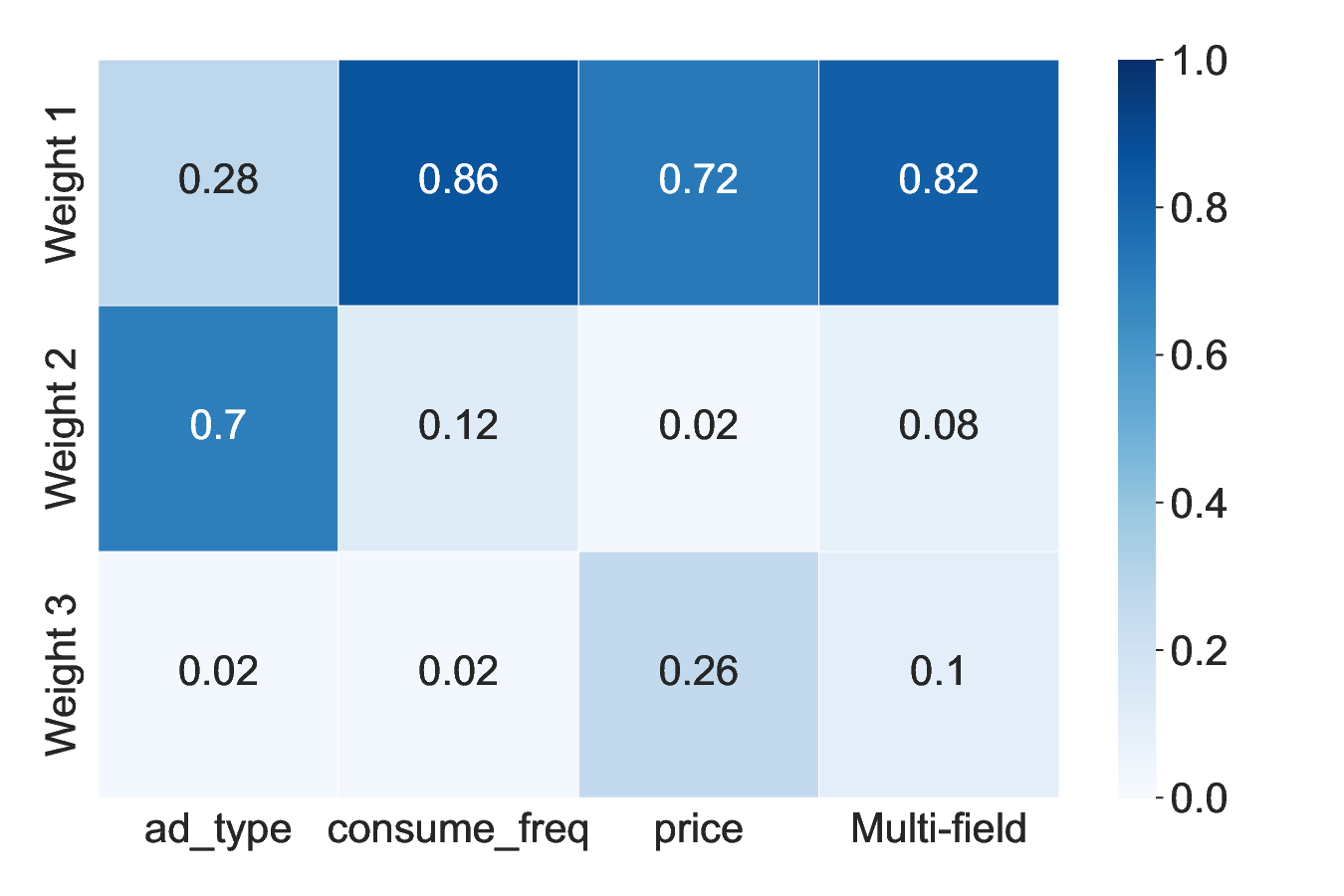}
    \vspace{-0.15in}

  \caption{Fusion weights in multi-field joint calibration.}
    \vspace{-0.2in}

  \label{fig.heatmap}
\end{figure}

\subsection{Analysis of Multi-field Fusion Weights}
The multi-field joint calibration of ConfCalib obtains fusion weights for multiple target fields by minimizing the calibration error on single or multiple fields.
Here, we select three target fields, "ad\_type", "consume\_frequency", and "price" in our industrial dataset, 
and optimize Field-RCE of each field and Multi-Field-RCE using grid search to acquire four groups of fusion weights separately, as shown in Fig.~\ref{fig.heatmap}. 
Each column takes Field-RCE of the corresponding field as the optimization object.
We can see that each field makes effects with different weights when optimizing Multi-Field-RCE of all three fields.
When optimizing Field-RCE on a single field, the weights of other fields are not zero, indicating that joint calibration on multiple fields can also reduce the single-field calibration error. 
The weight of "ad\_type" is generally the highest, suggesting that this field can better reflect global biases. 
Nevertheless, when Field-RCE on "ad\_type" is minimized, the weight of the other field "consume\_frequency" is the highest instead.
This indicates that calibration on a single field is insufficient, and multi-field calibration can provide effective assistance.

\subsection{Hyperparameter Analysis}
\begin{figure}[!t]
  \centering
    \vspace{-0.1in}
    \includegraphics[width=0.85\linewidth]{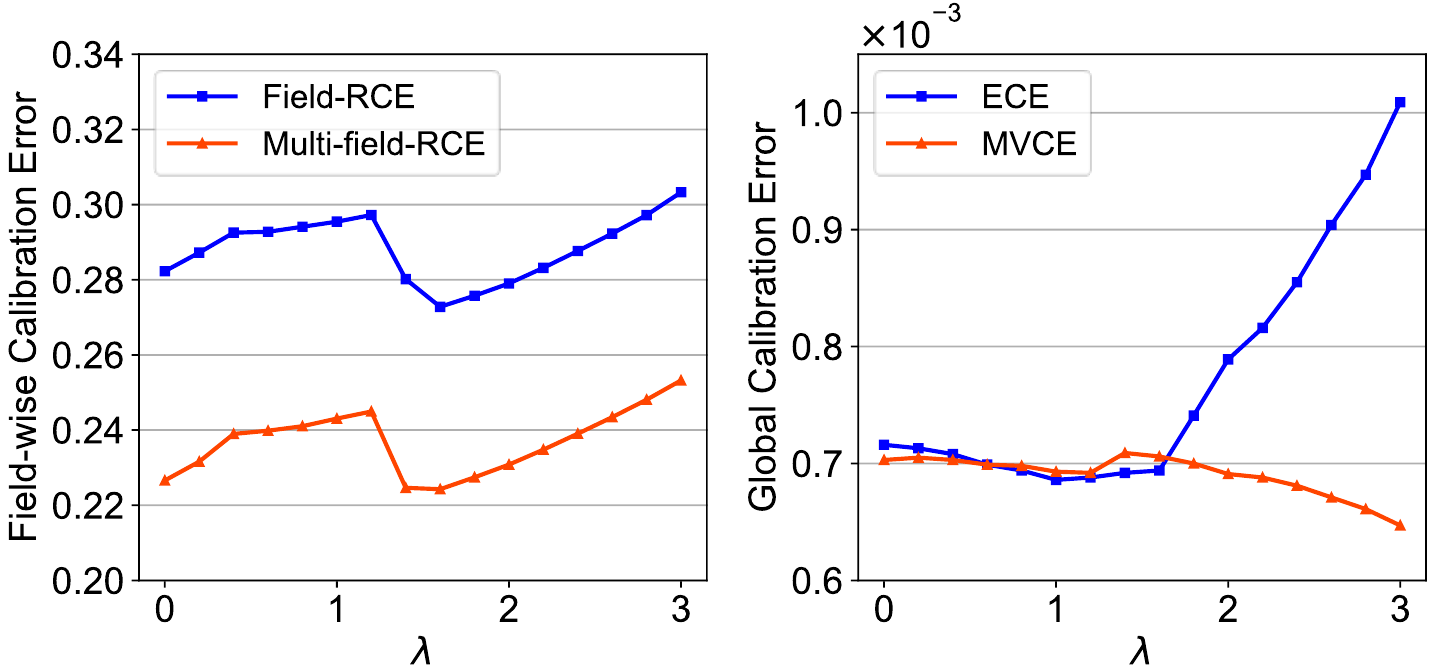}
    \vspace{-0.15in}
  \caption{Calibration errors w.r.t hyperparameter $\lambda$.}
    \vspace{-0.17in}
  \label{fig.hyperparams}
\end{figure}

Our method has a single hyperparameter $\lambda$ in the z-score mapping function Eq.~(\ref{eq.z_map}) to control the calibration strength manually. 
Its main effect is to select a relatively higher confidence level when observed samples are adequate, thus making the calibrated prediction closer to the observation. 
Here, we discuss the influence of different settings of $\lambda$. 
We set $\lambda=0,0.2,0.4\dots,3.0$ and experiment on the industrial dataset. 
Results are shown in Fig.~\ref{fig.hyperparams}.
Overall, the variation of field-wise calibration errors like Field-RCE and Multi-Field-RCE is within a small range as $\lambda$ changes. 
For global calibration error metrics, the change of ECE is relatively violent when $\lambda$ is significant, while MVCE remains almost the same. 
When $\lambda=0$, the awareness of confidence is removed, which means we disregard the model prediction and believe entirely in the observation.
We achieve a local optimum at this point.
As $\lambda$ increases, the calibration intensity gradually decreases, and the global optimal $\lambda$ is around 1.6.

\subsection{Online A/B Test}

We conduct one week's online A/B test separately on two industrial scenarios to evaluate ConfCalib's effectiveness in practical applications.
In each scenario, we randomly split users into two groups, each with tens of millions of users daily.
One group receives recommendations from ConfCalib, while the other gets recommendations from a well-crafted baseline.
On Huawei's online advertising platform, ConfCalib achieves a relative improvement of 2.42\% in CVR compared with the baseline.
On Top-grossing of Huawei AppGallery, ConfCalib obtains 
relative improvement of 32.6\% and 49.1\% in CTR and revenue. 
Top-grossing is a small-scale scenario where data distribution shifts frequently, and ConfCalib can better adapt this variability to achieve a significant improvement.

%% file: sections/conclusion.tex
\section{Conclusion}
In this paper, we propose a confidence-aware multi-field model calibration method, ConfCalib, to address the common phenomenon of data sparsity in model calibration for advertising recommendation. 
ConfCalib dynamically adjusts calibration intensity based on the confidence level obtained from the observed data distribution.
Moreover, it considers biases along multiple feature fields and introduces joint calibration to capture various bias patterns and further alleviate the impact of data sparsity. 
We conduct sufficient offline and online experiments.
Results demonstrate that ConfCalib outperforms prior methods and exhibits strong robustness against data sparsity. 
ConfCalib is currently running on Huawei's advertising platform and AppGallery and has achieved continuous online revenue improvement.